\documentclass[10pt,twocolumn,letterpaper]{article}
\usepackage{conf}

\usepackage{times}
\usepackage{epsfig, graphicx}
\usepackage{color}
\usepackage{xcolor}
\usepackage{amsmath,amssymb}
\usepackage[ruled, vlined, linesnumbered]{algorithm2e}
\usepackage{multirow}
\usepackage{array}
\usepackage{slashbox}
\usepackage{subfigure}
\usepackage{units}
\pdfoutput=1
\graphicspath{{./figures/}}
\DeclareGraphicsExtensions{.pdf,.jpeg,.png,.jpg}

\usepackage[pagebackref=true,breaklinks=true,letterpaper=true,colorlinks,bookmarks=false]{hyperref}

\usepackage{xspace}
\def\sota{state-of-the-art}

\newcommand{\refsec}[1]{Section~\ref{#1}}
\newcommand{\reffig}[1]{Figure~\ref{#1}}
\newcommand{\refeq}[1]{Equation~\eqref{#1}}

\newcommand{\reftab}[1]{Table~\ref{#1}}

\conffinalcopy 

\def\confPaperID{} 

\ifconffinal\fi
\begin{document}

\title{DeepFlux for Skeletons in the Wild}

\author{
Yukang Wang$^1$ \quad Yongchao Xu$^1$ \quad Stavros Tsogkas$^2$ \quad Xiang Bai$^1$ \quad Sven Dickinson$^2$ \quad Kaleem Siddiqi$^3$
\\[0.2cm]
$^1$Huazhong University of Science and Technology
\quad
$^2$University of Toronto
\\
$^3$School of Computer Science and Centre for Intelligent Machines, McGill University
}

\maketitle

\begin{abstract}


Computing object skeletons in natural images is challenging, owing to large variations in object appearance and scale, and the complexity of handling background clutter. Many recent methods frame object skeleton detection as a binary pixel classification problem, which is similar in spirit to learning-based edge detection, as well as to semantic segmentation methods. In the present article, we depart from this strategy by training a CNN to predict a two-dimensional vector field, which maps each scene point to a candidate skeleton pixel, in the spirit of flux-based skeletonization algorithms. This ``image context flux'' representation has two major advantages over previous approaches. First, it explicitly encodes the relative position of skeletal pixels to semantically meaningful entities, such as the image points in their spatial context, and hence also the implied object boundaries. Second, since the skeleton detection context is a region-based vector field, it is better able to cope with object parts of large width.
We evaluate the proposed method on three benchmark datasets for skeleton detection and two for symmetry detection, achieving consistently superior performance over state-of-the-art methods.
\end{abstract}

\section{Introduction} \label{sec:intro}

\begin{figure}
\centering
\subfigure[Previous CNN-based skeleton detections rely on NMS.]
{
\includegraphics[width=0.98\linewidth]{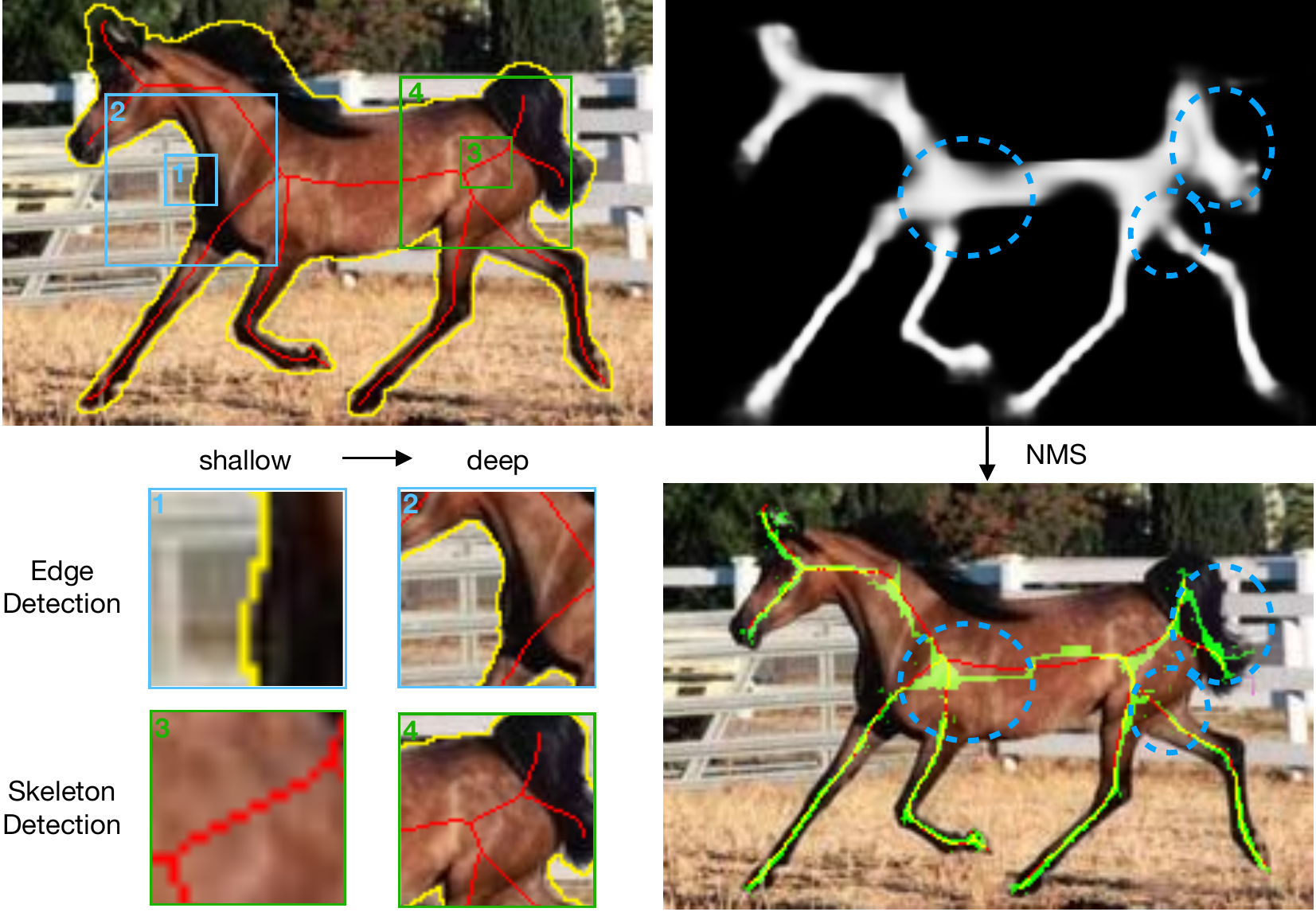}
\label{fig:motivation_a}
}
\subfigure[Flux provides an alternative way for accurately detecting skeletons.]
{
\includegraphics[width=0.98\linewidth]{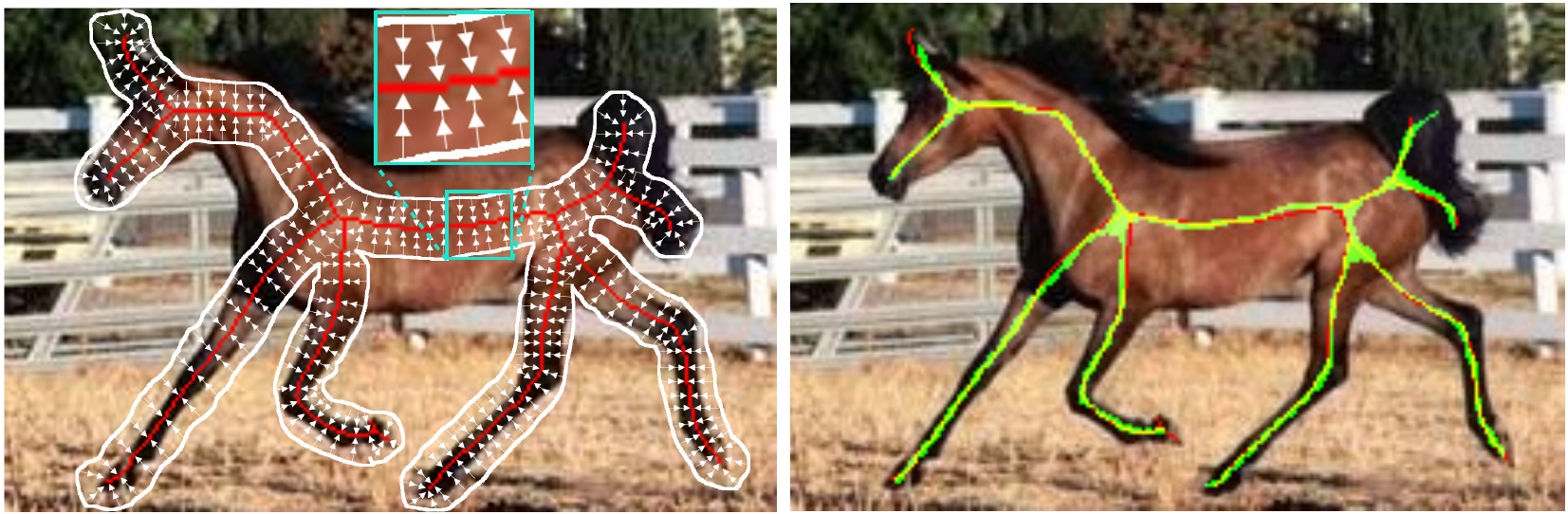}
\label{fig:motivation_b}
}
\caption{(a) Previous CNN-based methods treat skeleton detection as binary pixel classification, followed by non-maximum suppression (NMS). This can result in poor localization as well as poor connectedness. (b) The proposed DeepFlux method models skeleton context via a novel flux representation (left). The flux vector field encodes skeleton position in the context of the associated image pixels, and hence also the implied object boundaries. This allows one to associate skeletal pixels with sinks, where flux is absorbed, in the spirit of flux-based skeletonization methods \cite{siddiqi2002hamilton}. Red: ground truth skeleton; Green: detected skeleton.}
\label{fig:motivation}
\end{figure}

The shape skeleton, or medial axis~\cite{blum1973biological}, is a structure-based object descriptor that reveals local symmetry as well as connectivity between object parts~\cite{marr1978representation,dickinson2009object}. Modeling
objects via their axes of symmetry, and in particular, using skeletons,
has a long history in computer vision. Skeletonization algorithms
provide a concise and effective
representation of deformable objects, while supporting many applications,
including object recognition and retrieval~\cite{zhu1996forms,felzenszwalb2005pictorial,bai2009active,trinh2011skeleton}, pose estimation~\cite{girshick2011efficient,shotton2011real,wei2016convolutional}, hand gesture recognition~\cite{ren2013robust}, shape matching~\cite{siddiqi1999shock}, scene text detection~\cite{zhang2015symmetry},
and road detection in aerial scenes~\cite{sironi2014multiscale}.


Early algorithms for computing skeletons directly from images~\cite{lindeberg1998edge,liu1998segmenting,jang2001pseudo,yu2004segmentation,nedzved2006gray,zhang2007accurate,lindeberg2013scale} yield a gradient intensity map, driven by geometric constraints between skeletal pixels and edge fragments. Such methods cannot easily handle complex image data without prior information about object shape and location. Learning-based methods~\cite{levinshtein2013multiscale,sie2013detecting,tsogkas2012mil,shen2016misl,sironi2014multiscale} have an improved ability for object skeleton detection in natural images, but such methods are still unable to cope with complex backgrounds or clutter.

The recent surge of work in convolutional neural networks (CNNs) has lead to vast improvements in the performance of object skeleton detection algorithms~\cite{shen2016fsds,shen2017lmsds,ke2017srn,liu2017twostream,zhao2018hifi,liu2018lsn}. These existing CNN-based methods usually derive from Holistically-Nested Edge Detection (HED)~\cite{xie2015hed}, and frame the problem as binary pixel classification.
Most such approaches focus on designing an appropriate network and leveraging better multi-level features for capturing skeletons across a range of spatial scales.



Object skeleton computation using CNNs from natural images is inherently different from the problem of edge detection. As illustrated in~\reffig{fig:motivation_a}, edges associated with object boundaries can typically be detected locally, due to a local appearance change or a change in texture. Thus, the shallow convolutional layers, with accurate spatial information, can capture potential edge locations. Object skeletons, though, have to do with medial properties and high-level semantics. In particular, skeletons are situated at regions within object parts, where there is a local symmetry, since the medial axis bisects the object angle \cite{siddiqi2008}. Capturing this purely from local image information (\eg, the green box numbered 3 in~\reffig{fig:motivation_a}) is not feasible, since this requires a larger spatial extent, in this case the width of the torso of the horse. Since shallow layers do not allow skeletal points to be captured, deeper layers of CNNs, with associated coarser features, are required.
But this presents a confound -- such coarse features may not provide accurate spatial localization of the object skeleton.


\begin{figure*}
\centering
\includegraphics[width=0.90\linewidth]{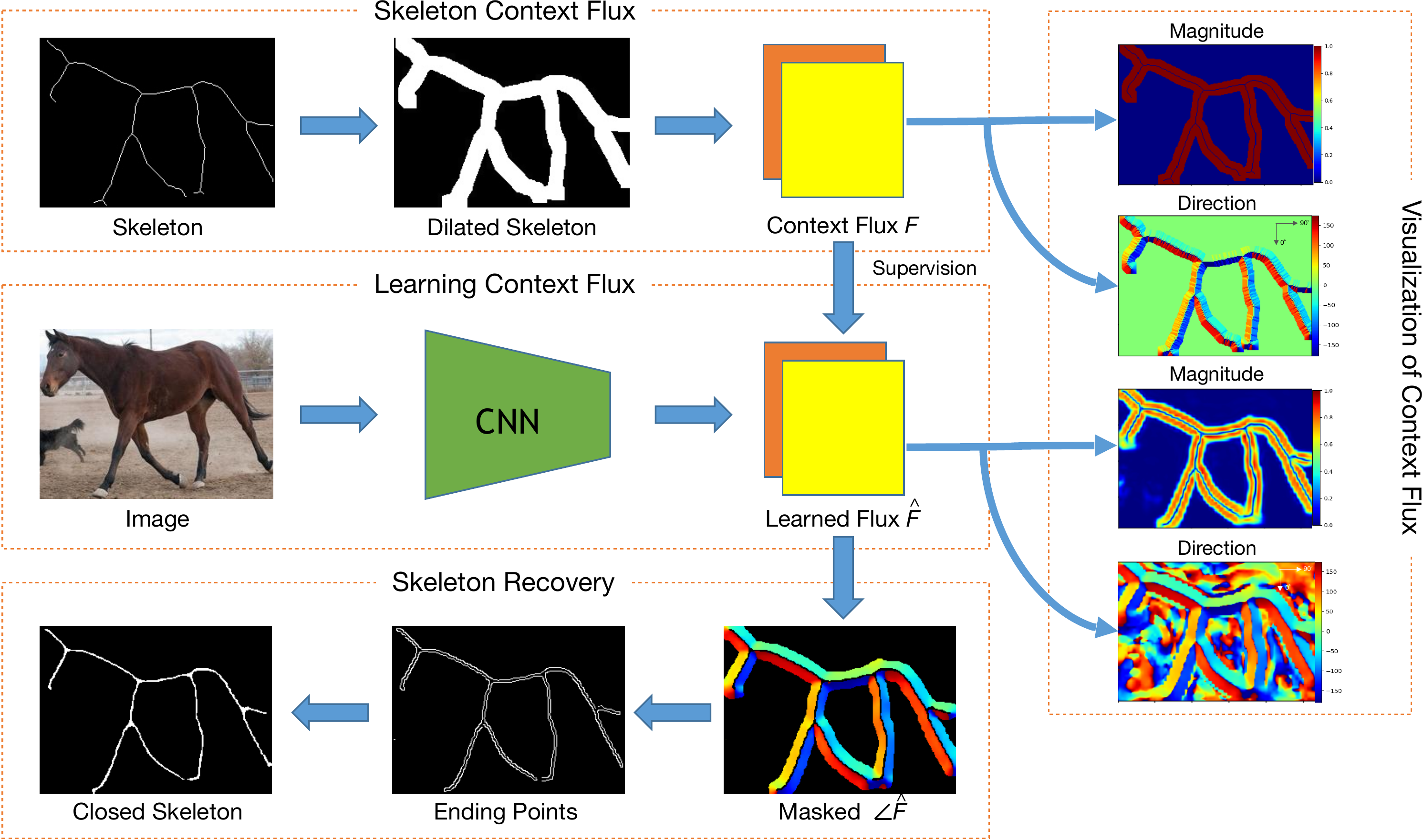}
\vskip 0.4cm
\caption{The DeepFlux pipeline. Given an input image, the network computes a two-dimensional vector field of skeleton context flux (visualizations of magnitude and direction on the right). The object skeleton is then recovered by localizing points where the net inward flux is high, followed by a morphological closing operation.}
\label{fig:pipeline}
\end{figure*}


In this paper, we propose a novel notion of image context flux, to accurately detect object skeletons within a CNN framework. More precisely, we make use of skeleton context by using a two-dimensional vector field to capture a flux representation. For each skeleton context pixel, the flux is defined by the two-dimensional unit vector pointing to its nearest skeleton pixel. Within this flux representation, the object skeleton corresponds to pixels where the net inward flux is positive, following the motivation behind past flux-based methods for skeletonizing binary objects \cite{siddiqi2002hamilton,dimitrov2003}. We then develop a simple network to learn the image context flux, via a pixel-wise regression task in place of binary classification. Guided by the learned context flux encoding the relative location between context pixels and the skeleton, we can easily and accurately recover the object skeleton. In addition, the skeleton context provides a larger receptive field size for estimation, which is potentially helpful for detecting skeletons associated with larger spatial scales.

The main contributions of this paper are three-fold:
1) We propose a novel \emph{context flux} to represent the object skeleton. This concept explicitly encodes the relationship between image pixels and their closest skeletal points.
2) Based on the context flux, we develop a method which we dub DeepFlux, that accurately and efficiently detects object skeletons in an image.
3) DeepFlux consistently outperforms \sota\ methods on five public benchmarks. To our knowledge this is the first application of flux concepts, which have been successfully used for skeletonization of binary objects, to the detection of object skeletons in natural images. It is also the first attempt at learning such flux-based representations directly from natural images.

The rest of this paper is organized as follows. We review related work in~\refsec{sec:related}. We develop the DeepFlux method in~\refsec{sec:method} and carry out an extensive experimental evaluation in~\refsec{sec:experiments}. We then conclude with a discussion of our results in~\refsec{sec:conclusion}.

\section{Related Work} \label{sec:related}

Object skeletonization has been widely studied in recent decades. In our review, we contrast traditional methods with those based on deep learning.
\medskip

\noindent\textbf{Traditional methods:}
Many early skeleton detection algorithms~\cite{lindeberg1998edge,liu1998segmenting,jang2001pseudo,yu2004segmentation,nedzved2006gray,zhang2007accurate,lindeberg2013scale} are based on gradient intensity maps. In~\cite{siddiqi2002hamilton}, the authors study the limiting average outward flux of the gradient of a Euclidean distance function to a 2D or 3D object boundary. The skeleton is associated with those locations where an energy principle is violated, where there is a net inward flux. Other researchers have constructed the skeleton by merging local skeleton segments with a learned segment-linking model. Levinshtein {\em et al.}~\cite{levinshtein2013multiscale} propose a method to work directly on images, which uses multi-scale super-pixels and a learned affinity between adjacent super-pixels to group proximal medial points. A graph-based clustering algorithm is then applied to form the complete skeleton. Lee {\em et al.}~\cite{sie2013detecting} improve the approach in~\cite{levinshtein2013multiscale} by using a deformable disc model, which can detect curved and tapered symmetric parts. A novel definition of an appearance medial axis transform (AMAT) has been proposed in~\cite{tsogkas2017amat}, to detect symmetry in the wild in a purely bottom up, unsupervised fashion. In~\cite{jerripothula2017object}, the authors present an unconventional method based on joint co-skeletonization and co-segmentation.

In other literature~\cite{tsogkas2012mil,shen2016misl,sironi2014multiscale}, object skeleton detection is treated as a pixel-wise classification or regression problem. Tsogkas and Kokkinos~\cite{tsogkas2012mil} extract hand-designed features at each pixel and train a classifier for symmetry detection. They employ a multiple instance learning (MIL) framework to accommodate for the unknown scale and orientation of symmetry axes. Shen {\em et al.}~\cite{shen2016misl} extend the approach in~\cite{tsogkas2012mil} by training a group of MIL classifiers to capture the diversity of symmetry patterns. Sironi {\em et al.}~\cite{sironi2014multiscale} propose a regression-based approach to improve the accuracy of skeleton locations. They train regressors which learn the distances to the closest skeleton in scale-space and identify the skeleton by finding the local maxima.

\medskip

\noindent\textbf{Deep learning-based methods:}
With the popularization of CNNs, deep learning-based methods~\cite{shen2016fsds,shen2017lmsds,ke2017srn,liu2017twostream,zhao2018hifi,liu2018lsn} have had a tremendous impact on object skeleton detection. Shen {\em et al.}~\cite{shen2016fsds} fuse scale-associated deep side-outputs (FSDS) based on the architecture of HED~\cite{xie2015hed}. Given that the skeleton of different scales can be captured in different stages, they supervise the side outputs with scale-associated ground-truth data. Shen {\em et al.}~\cite{shen2017lmsds} then extend their original method by learning multi-task scale-associated deep side outputs (LMSDS).
This leads to improved skeleton localization, scale prediction, and better overall performance. Ke {\em et al.}~\cite{ke2017srn} present a side-output residual network (SRN), which leverages the output residual units to fit the errors between the ground-truth and the side-outputs. By cascading residual units in a deep-to-shallow manner, SRN can effectively detect the skeleton at different scales. Liu {\em et at.}~\cite{liu2017twostream} develop a two-stream network that combines image and segmentation cues to capture complementary information for skeleton localization. In~\cite{zhao2018hifi}, the authors introduce a hierarchical feature integration (Hi-Fi) mechanism. By hierarchically integrating multi-scale features with bidirectional guidance, high-level semantics and low-level details can benefit from each other. Liu {\em et al.}~\cite{liu2018lsn} propose a linear span network (LSN) that uses linear span units to increase the independence of convolutional features and the efficiency of feature integration.

\medskip
Though the method we propose in the present paper benefits from CNN-based learning, it differs from the methods in  ~\cite{shen2016fsds,shen2017lmsds,ke2017srn,liu2017twostream,zhao2018hifi,liu2018lsn} in a fundamental way, due to its different learning objective. Instead of treating object skeleton detection in natural images as a binary classification problem, DeepFlux focuses on learning the context flux of skeletons, and as such includes more informative non-local cues, such as the relative position of skeleton points to image points in their vicinity, and thus also, implicitly, the associated object boundaries. A direct consequence of this powerful image context flux representation is that a simple post-processing step can recover the skeleton directly from the learned flux, avoiding inaccurate localizations of skeletal points caused by non-maximum suppression in previous deep learning methods.
In addition, DeepFlux enlarges the spatial extent used by the CNN to detect the skeleton, through the use of skeleton context flux. This allows our approach to capture larger object parts.

We note that the proposed DeepFlux is in spirit similar with the original notion of flux \cite{siddiqi2002hamilton,dimitrov2003} that is defined based on an object boundary, for skeletonization of 2D/3D binary objects. As such, DeepFlux inherits its mathematical properties including the unique mapping of skeletal points to boundary points. However, we are the first to extend this notion of flux to skeleton detection in natural images, by computing the flux on dilated skeletons for supervised learning. Our work is also related to the approaches in ~\cite{bai2017deep,maninis2018convolutional} which learn the direction cues for edge detection and instance segmentation. In the present article, this direction information is encoded in the flux representation, and is implicitly learned for skeleton recovery.

\section{Method} \label{sec:method}
Many recent CNN-based skeleton detection approaches build on some variant of the HED architecture~\cite{xie2015hed}.
The combination of a powerful classifier (CNN) and the use of side outputs to extract and combine features at multiple scales has enabled these systems to accurately localize medial points of objects in natural images.
However, while \sota\ skeleton detection systems are quite effective at extracting medial axes of elongated structures, they still struggle when reasoning about
ligature areas.
This is expected: contrary to the skeletal branches they connect, ligature areas exhibit much less structural regularity, making their exact localization ambiguous.
As a result, most methods result in poor localization of ligature points, or poor connectedness between medial axes of object parts.

We propose to remedy this issue by casting skeleton detection as the problem of predicting a two-dimensional flux field from scene points to nearby skeleton points, within a fixed-size neighborhood.
We then define skeleton points as the local flux minima, or, alternatively, as sinks ``absorbing'' flux from nearby points.
We argue --and prove empirically in our experiments-- that this approach leads to more robust localization and better connectivity between skeletal branches.
We also argue that considering a small neighborhood around the true skeleton points is sufficient, consistent with
past approaches to binary object skeletonization \cite{dimitrov2003}.
Whereas predicting the flux for the entire object would allow us to also infer the medial radius function, in this work we focus on improving medial point localization. The overall pipeline of the proposed method, aptly named \emph{DeepFlux}, is depicted in~\reffig{fig:pipeline}.


\begin{figure}
\centering
\includegraphics[width=0.98\linewidth]{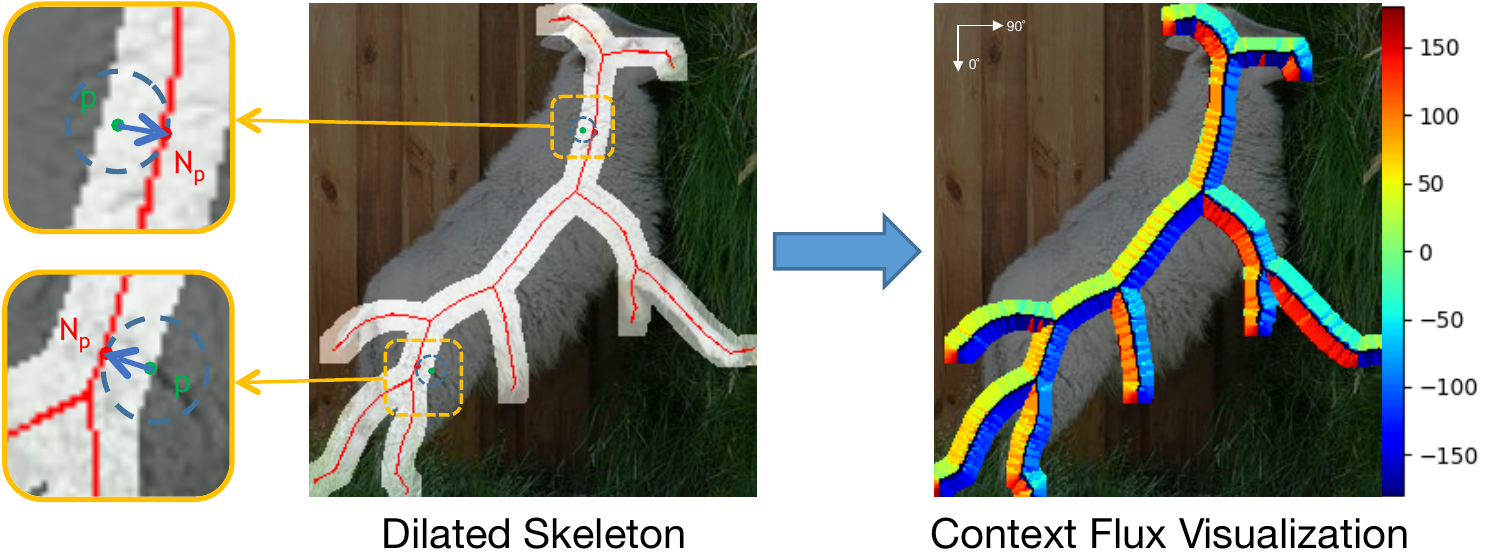}
\vskip 0.2cm
\caption{For each context (non-skeleton) pixel $p$ in the dilated skeleton mask, we find its nearest skeleton pixel $N_p$. The flux $F(p)$ is defined as the two-dimensional unit vector that points away from $p$ to $N_p$. For skeleton points, the flux is set to $(0, 0)$. On the right, we visualize the direction of the flux field.}
\label{fig:directionfield}
\end{figure}

\subsection{Skeleton context flux} \label{sec:directionfield}
We represent $F(x,y) = (F_x, F_y)$ as a two-channel map with continuous values corresponding to the x and y coordinates of the flux vector respectively.
An intuitive visualization is shown in~\reffig{fig:directionfield}.
When skeleton detection is framed as a binary classification task, ground truth is a 1-pixel wide binary skeleton map; for our \emph{regression} problem the ground truth must be modified appropriately.

We divide a binary skeleton map into three non-overlapping regions: 1) {\it skeleton context}, $R_c$, which is a the vicinity of the skeleton; 2) {\it skeleton pixels}, denoted by $R_s$; and 3) {\it background pixels}, $R_b$. In practice, $R_c$ is obtained by dilating the binary skeleton map with a disk of radius $r$, and subtracting skeleton pixels $R_s$. Then, for each context pixel $p \in R_c$, we find its nearest skeleton pixel $N_p \in R_s$. A unit direction vector that points away from $p$ to $N_p$ is then computed as the flux on the context pixel $p$.
This can be efficiently computed with the aid of a distance transform algorithm.\footnote{In fact, in the context of skeletonization of binary objects \cite{siddiqi2008}, this flux vector would be in the direction opposite to that of the spoke vector from a skeletal pixel to its associated boundary pixel.} For the remaining pixels composed of $R_s$ and $R_b$, we set the flux to $(0, 0)$. Formally, we have:
\begin{equation}
F(p) =
\left\{
\begin{matrix} \
\overrightarrow{p N_p}/\left\vert\overrightarrow{p N_p}\right\vert, & p\in R_c  \\ \\
(0,0), & p \in R_s \cup R_b,
\end{matrix}
\right.
\label{eq:sklrep}
\end{equation}
where $\left\vert\overrightarrow{p N_p}\right\vert$ denotes the length of the vector from $p$ to $N_p$.

As a representation of the spatial context associated with each skeletal pixel, our proposed image context flux possesses a few distinct advantages when used to detect object skeletons in the wild.
Unlike most learning approaches that predict skeleton probabilities individually for each pixel, our DeepFlux method leverages consistency between flux predictions within a neighborhood around each candidate pixel.
Conversely, if the true skeleton location changes, the surrounding flux field will also change noticeably.
A beneficial side-effect is that our method does not rely directly on the coarse responses produced by deeper CNN layers for localizing skeletons at larger scales, which further reduces localization errors.
As we show in our experiments, these properties make our method more robust to the localization of skeleton points, especially around ligature regions, and less prone to gaps, discontinuities, and irregularities caused by local mispredictions.
Finally, it is easy to accurately recover a binary object skeleton using the magnitude and direction of the predicted flux, as explained in~\refsec{sec:postprocessing}.


\begin{figure}
\centering
\includegraphics[width=0.93\linewidth]{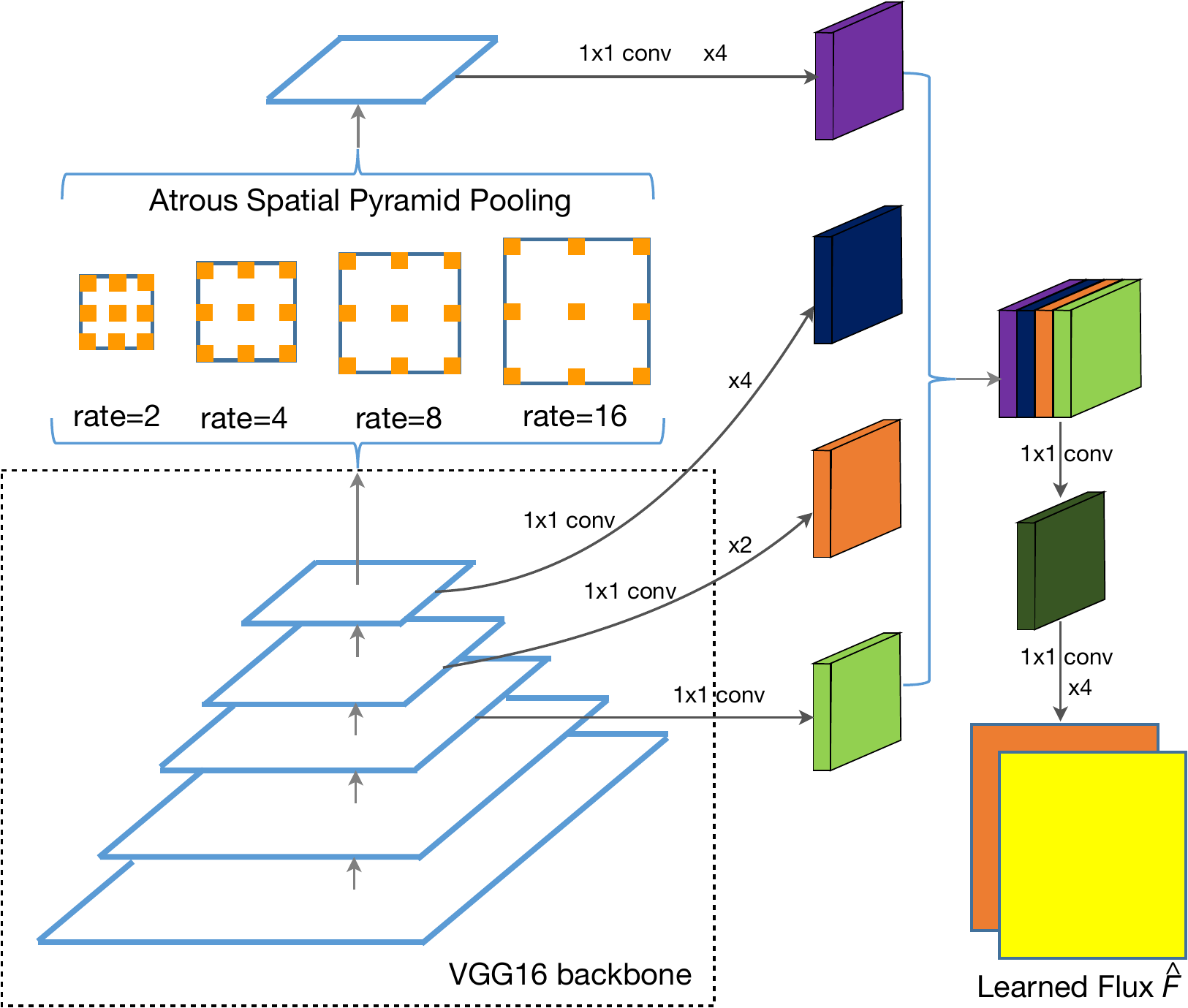}
\vskip 0.2cm
\caption{Network architecture. We adopt the pre-trained VGG16~\cite{vgg16network} with the ASPP module~\cite{chen2018deeplab} as the backbone network and with multi-level feature fusion via concatenation. The network is trained to predict the proposed context flux $F$, which is an image representing a two-dimensional vector field.}
\label{fig:networkarchitecture}
\end{figure}

\subsection{Network architecture} \label{sec:networkarchitecture}
The network for learning the skeleton context flux follows  closely the fully convolutional architecture of~\cite{long2015fcn}, and is shown in~\reffig{fig:networkarchitecture}.
It consists of three modules:
1) a backbone network used to extract 3D feature maps;
2) an ``atrous'' spatial pyramid pooling (ASPP) module~\cite{chen2018deeplab} to enlarge the receptive field while avoiding excessive downsampling; and
3) a multi-stage feature fusion module.

To ensure a fair comparison with previous work, we also adopt VGG16~\cite{vgg16network} as the backbone network.
As in~\cite{xie2015hed}, we discard the last pooling layer and the fully connected layers that follow.
The use of the atrous module is motivated by the need for a wide receptive field: when extracting skeletons we have to guarantee that the receptive field of the network is wider than the largest medial radius of an object part in the input image.
The receptive field of the VGG16 backbone is 196, which is not wide enough for large objects.
Furthermore, it has been demonstrated in~\cite{luo2016receptivefield} that the effective receptive field only takes up a fraction of the full theoretical receptive field.
Thus, we employ ASPP to capture multi-scale information.
Specifically, four parallel atrous convolutional layers with $3\times3$ kernels but different atrous rates (2, 4, 8, 16) are added to the last layer of the backbone, followed by a concatenation along the channel dimension.
In this way, we obtain feature maps with a theoretical receptive field size of 708 which we have found to be large enough for the images we have experimented on.

To construct a multi-scale representation of the input image, we fuse the feature maps from side outputs at conv3, conv4, conv5, and ASPP layers, after convolving them with a  $1 \times 1$ kernel.
Since feature maps at different levels have different spatial resolutions, we resize them all to the dimensions of conv3 before concatenating them.
Prediction is then performed on the fused feature map, and then up-sampled to the dimensions of the input image.
For up-sampling we use bilinear interpolation.
The final output of the network is a 2-channel response map containing predictions of the x and y coordinates of the image content flux field $\hat{F}(p)$ for every pixel $p$ in the image.

\subsection{Training objective} \label{sec:trainingobjective}
We choose the $L_2$ loss function as our training objective.
Due to a severe imbalance in the number of context and background pixels, we adopt a class-balancing strategy similar to the one in~\cite{xie2015hed}.
Our balanced loss function is
\begin{equation}
L \; = \; \sum_{p \in \Omega}{w(p) * \left\|F(p), \hat{F}(p)\right\|_2},
\label{eq:loss}
\end{equation}
where $\Omega$ is the image domain, $\hat{F}(p)$ is the predicted flux, and $w(p)$ denotes the weight coefficient of pixel $p$.
The weight $w(p)$ is calculated as follows:
\begin{equation}
w(p) =
\left\{
\begin{matrix} \
\frac{|R_b|}{|R_c|+|R_b|+|R_s|}, & p \in R_c \cup R_s \\ \\
\frac{|R_c|+|R_s|}{|R_c|+|R_b|+|R_s|}, & p \in R_b,
\end{matrix}
\right.
\label{eq:weight}
\end{equation}
where $|R_c|$, $|R_b|$ and $|R_s|$ denote the number of context, background, and skeleton pixels, respectively.

\subsection{From flux to skeleton points} \label{sec:postprocessing}
We propose a simple post-processing procedure to recover an object skeleton from the predicted context flux.
As described in~\refeq{eq:sklrep}, pixels around the skeleton are labeled with unit two-dimensional vectors while the others are set to $(0, 0)$.
Thus, thresholding the magnitude of the vector field reveals the context pixels while
computing the flux direction reveals the location of context pixels relative to the skeleton. We refer the reader to~\reffig{fig:pipeline} for a visualization of the post-processing steps, listed in Algorithm~\ref{algo:skeletonrecovery}.

Let $|\hat{F}|$ and $\angle \hat{F}$ be the magnitude and direction of the predicted context flux $\hat{F}$, respectively.
For a given pixel $p$, $\angle \hat{F}(p)$ is binned into one of 8 directions, pointing to one of the 8 neighbors, denoted by $\mathcal{N}_{\angle \hat{F}(p)}(p)$.
Having computed these two quantities, extracting the skeleton is straightforward:
pixels close to the real object skeleton should have a high inward flux, due to a singularity in the vector field $\hat{F}$, as analyzed in \cite{dimitrov2003}.
Finally, we apply a morphological dilation with a disk structuring element of radius $k_1$, followed by a morphological erosion with a disk of radius $k_2$, to group pixels together and produce the object skeleton.

\begin{algorithm}[t]
\caption{Algorithm for skeleton recovery from learned context flux $\hat{F}$. $|\hat{F}|$: magnitude; $\angle \hat{F}$: direction; $\mathcal{N}_{\angle \hat{F}(p)}(p)$: neighbor of $p$ at direction $\angle \hat{F(p)}$.}
\label{algo:skeletonrecovery}
\DontPrintSemicolon

\KwIn{Predicted context flux $\hat{F}$, threshold $\lambda$}
\KwOut{Binary skeleton map $S$}
\SetKwBlock{Begin}{function}{end function}
\Begin($\text{Skeleton\_Recovery} {(} \hat{F}, \lambda {)}$)
{
    // initialization \\
    $S \gets \textbf{False}$ \\
    // find ending points near skeleton \\
    \ForEach{$p \in \Omega$}
    {
    	\If{$|\hat{F}(p)| > \lambda$ \textbf{and} $|\hat{F}(\mathcal{N}_{\angle \hat{F}(p)}(p))| \leq \lambda$}
        {
        	$S(p) \gets \textbf{True}$
        }
    }
    // apply morphological closing \\
    $S \gets \varepsilon_{k_2}(\delta_{k_1}(S))$ \;
    \Return{$S$} \;
}
\end{algorithm}

\section{Experiments} \label{sec:experiments}
We conduct experiments on five well-known, challenging datasets, including three for skeleton detection (SK-LARGE~\cite{shen2017lmsds}, SK506~\cite{shen2016fsds}, WH-SYMMAX~\cite{shen2016misl}) and two for local symmetry detection (SYM-PASCAL~\cite{ke2017srn}, SYMMAX300~\cite{tsogkas2012mil}).
We distinguish between the two tasks by associating skeletons with a foreground object, and local symmetry detection with any symmetric structure, be it a foreground object or background clutter.

\begin{table*}
\begin{center}
\begin{tabular}{|l|c|c|c|c|c|}
\hline
Methods      & SK-LARGE  & SK506  & WH-SYMMAX  &  SYM-PASCAL & SYMMAX300    \\ \hline\hline
MIL~\cite{tsogkas2012mil}  & 0.353   & 0.392  & 0.365 & 0.174 & 0.362     \\ \hline
HED~\cite{xie2015hed}  & 0.497   & 0.541  & 0.732 & 0.369 & 0.427    \\ \hline
RCF~\cite{liu2017rcf}  & 0.626   & 0.613  & 0.751 & 0.392 & -    \\ \hline
FSDS*~\cite{shen2016fsds}  & 0.633   & 0.623  & 0.769 & 0.418 & 0.467    \\ \hline
LMSDS*~\cite{shen2017lmsds} & 0.649 & 0.621 & 0.779 & - & -
\\ \hline
SRN~\cite{ke2017srn} & 0.678   & 0.632  & 0.780 & 0.443 & 0.446    \\ \hline
LSN~\cite{liu2018lsn} & 0.668   & 0.633  & 0.797 & 0.425 & 0.480    \\ \hline
Hi-Fi*~\cite{zhao2018hifi} & 0.724   & 0.681  & 0.805 & 0.454 & -    \\ \hline
DeepFlux (Ours) & \bf{0.732}  & \bf{0.695}  & \bf{0.840}  & \bf{0.502} & \bf{0.491} \\ \hline
\end{tabular}
\end{center}
\caption{F-measure comparison. * indicates scale supervision was also used. Results for competing methods are from the respective papers.}
\label{tab:quantitativeresults}
\end{table*}

\subsection{Dataset and evaluation protocol}\label{sec:datasets}
\noindent\textbf{SK-LARGE}~\cite{shen2017lmsds} is a benchmark dataset for object skeleton detection, built on the MS COCO dataset~\cite{chen2015mscoco}.
It contains 1491 images, 746 for training and 745 for testing.

\medskip

\noindent\textbf{SK506}~\cite{shen2016fsds} (aka SK-SMALL), is an earlier version of SK-LARGE containing 300 train images and 206 test images.

\medskip

\noindent\textbf{WH-SYMMAX}~\cite{shen2016misl} contains 328 cropped images from the Weizmann Horse dataset~\cite{borenstein2002horse}, with skeleton annotations.
It is split into 228 train images and 100 test images.

\medskip

\noindent\textbf{SYM-PASCAL}~\cite{ke2017srn} is derived from the PASCAL-VOC-2011 segmentation dataset~\cite{everingham2010pascalvoc} and targets object symmetry detection in the wild.
It consists of 648 train images and 787 test images.

\medskip

\noindent\textbf{SYMMAX300}~\cite{tsogkas2012mil} is built on the Berkeley Segmentation Dataset (BSDS300)~\cite{martin2001bsds}, which contains 200 train images and 100 test images.
Both foreground and background symmetries are considered.

\medskip

\noindent\textbf{Evaluation protocol}
We use precision-recall (PR) curves and the F-measure metric to evaluate skeleton detection performance in our experiments.
For methods that output a skeleton probability map, a standard non-maximal suppression (NMS) algorithm~\cite{dollar2015nms} is first applied and the thinned skeleton map is obtained.
This map is then thresholded into a binary map and matched with the groundtruth skeleton map, allowing small localization errors.
Since DeepFlux does not directly output skeleton probabilities, we use the inverse magnitude of predicted context flux on the recovered skeleton as a surrogate for a ``skeleton confidence''.
Thresholding at different values gives rise to a PR curve and the optimal threshold is selected as the one producing the highest F-measure according to the formula $F = 2PR/(P+R)$.
F-measure is commonly reported as a single scalar performance index.

\subsection{Implementation details} \label{sec:implementation}
Our implementation involves the following hyperparameters (values in parentheses denote the default values used in our experiments):
the width of the skeleton context neighborhood $r=7$;
the threshold used to recover skeleton points from the predicted flux field, $\lambda=0.4$;
the sizes of the structuring elements involved in the morphological operations for skeleton recovery, $k_1=3$ and $k_2=4$.

For training, we adopt standard data augmentation strategies~\cite{shen2016fsds,shen2017lmsds,zhao2018hifi}.
We resize training images to 3 different scales (0.8, 1, 1.2) and then rotate them to 4 angles ($0^\circ$, $90^\circ$, $180^\circ$, $270^\circ$).
Finally, we flip them with respect to different axes (up-down, left-right, no flip).
The proposed network is initialized with the VGG16 model pretrained on ImageNet~\cite{imagenet} and optimized using ADAM~\cite{kinga2015adam}.
The learning rate is set to $10^{-4}$ for the first 100k iterations, then reduced to $10^{-5}$ for the remaining 40k iterations.

We use the Caffe~\cite{jia2014caffe} platform to train DeepFlux. All experiments are carried out on a workstation with an Intel Xeon 16-core CPU (3.5GHz), 64GB RAM, and a single Titan Xp GPU.
Training on SK-LARGE using a batch size of 1 takes about 2 hours.



\subsection{Results} \label{sec:results}
\begin{figure} \centering
\subfigure[SK-LARGE]
{
\label{fig:prsklarge}
\includegraphics[width=0.46\columnwidth]{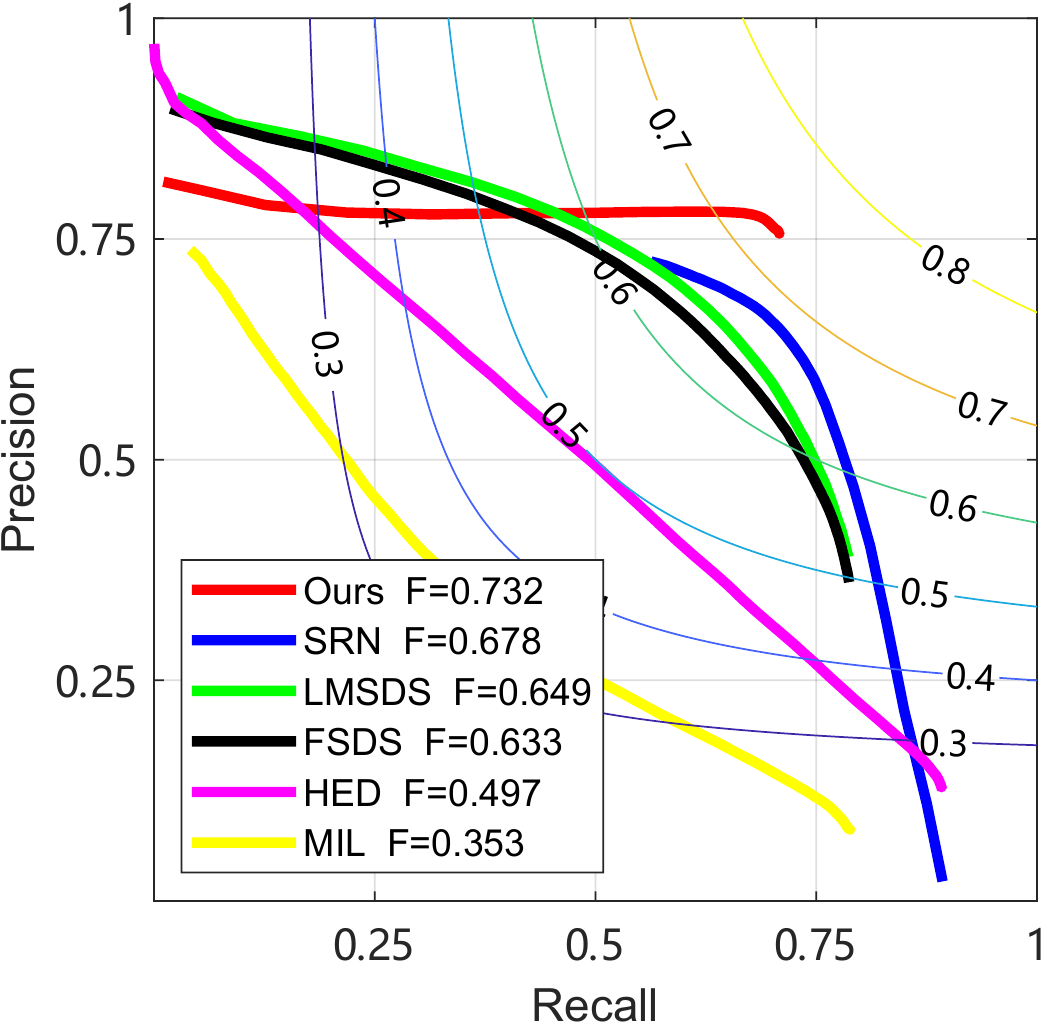}
}
\subfigure[SK506]
{
\label{fig:prsk506}
\includegraphics[width=0.46\columnwidth]{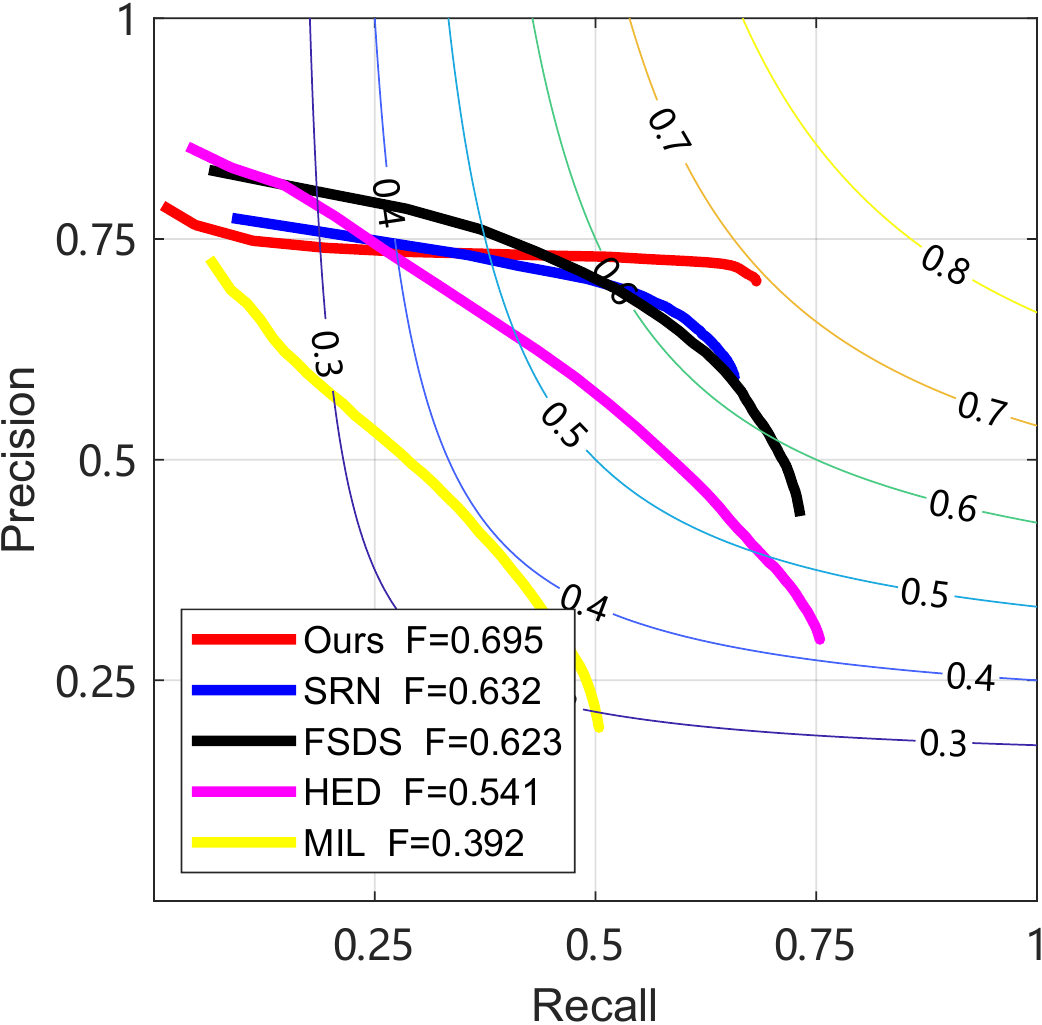}
}
\subfigure[WH-SYMMAX]
{
\label{fig:prsymmax}
\includegraphics[width=0.46\columnwidth]{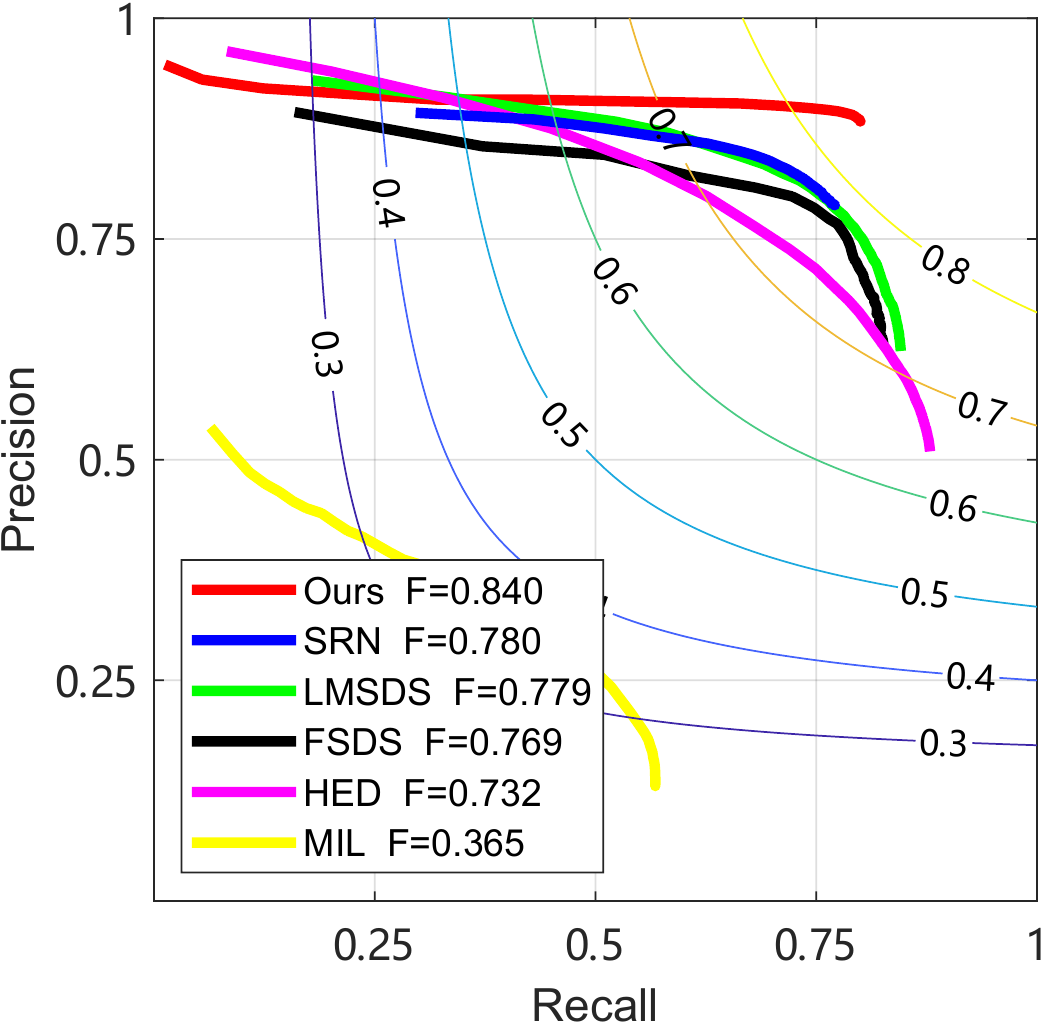}
}
\subfigure[SYM-PASCAL]
{
\label{fig:prsympascal}
\includegraphics[width=0.46\columnwidth]{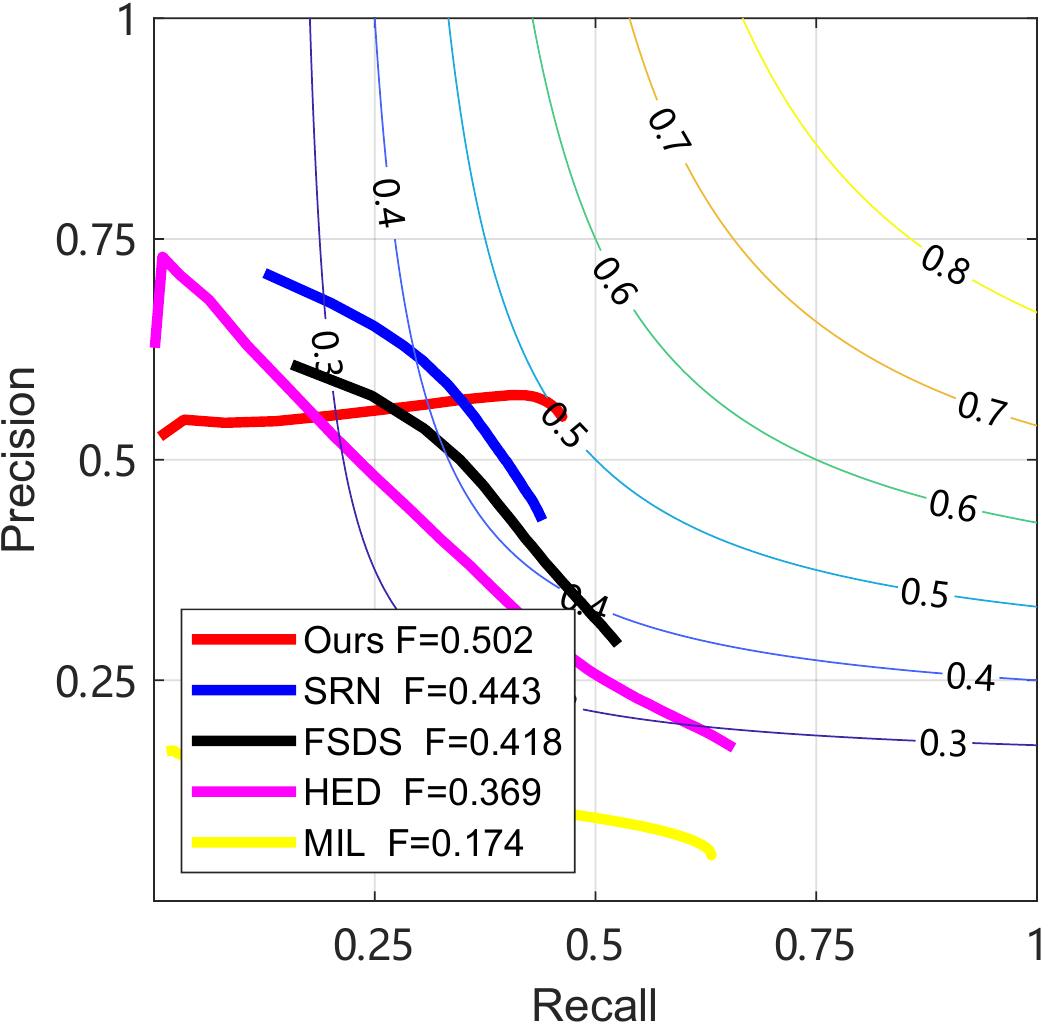}
}
\caption{PR curves on four datasets. DeepFlux offers high precision, especially in the high-recall regime.}
\label{fig:prcurve}
\end{figure}

PR-curves for all methods are shown in~\reffig{fig:prcurve}.
DeepFlux performance excels particularly in the high-precision regime, where it clearly surpasses competing methods.
This is indicative of the contribution of local context to more robust and accurate localization of skeleton points.

\reftab{tab:quantitativeresults} lists the optimal F-measure score for all methods.
DeepFlux consistently outperforms all other approaches using the VGG16 backbone~\cite{vgg16network}.
Specifically, it improves over the recent Hi-Fi~\cite{zhao2018hifi} by $0.8\%$, $1.4\%$, and $3.5\%$ on SK-LARGE, SK506, and WH-SYMMAX, respectively, despite the fact that Hi-Fi uses stronger supervision during training (skeleton position \emph{and} scale).
DeepFlux also outperforms LSN~\cite{liu2018lsn}, another recent method, by $6.4\%$, $6.2\%$, and $4.3\%$ on SK-LARGE, SK506, and WH-SYMMAX, respectively.

\begin{figure*}
\centering
\includegraphics[width=1.0\linewidth]{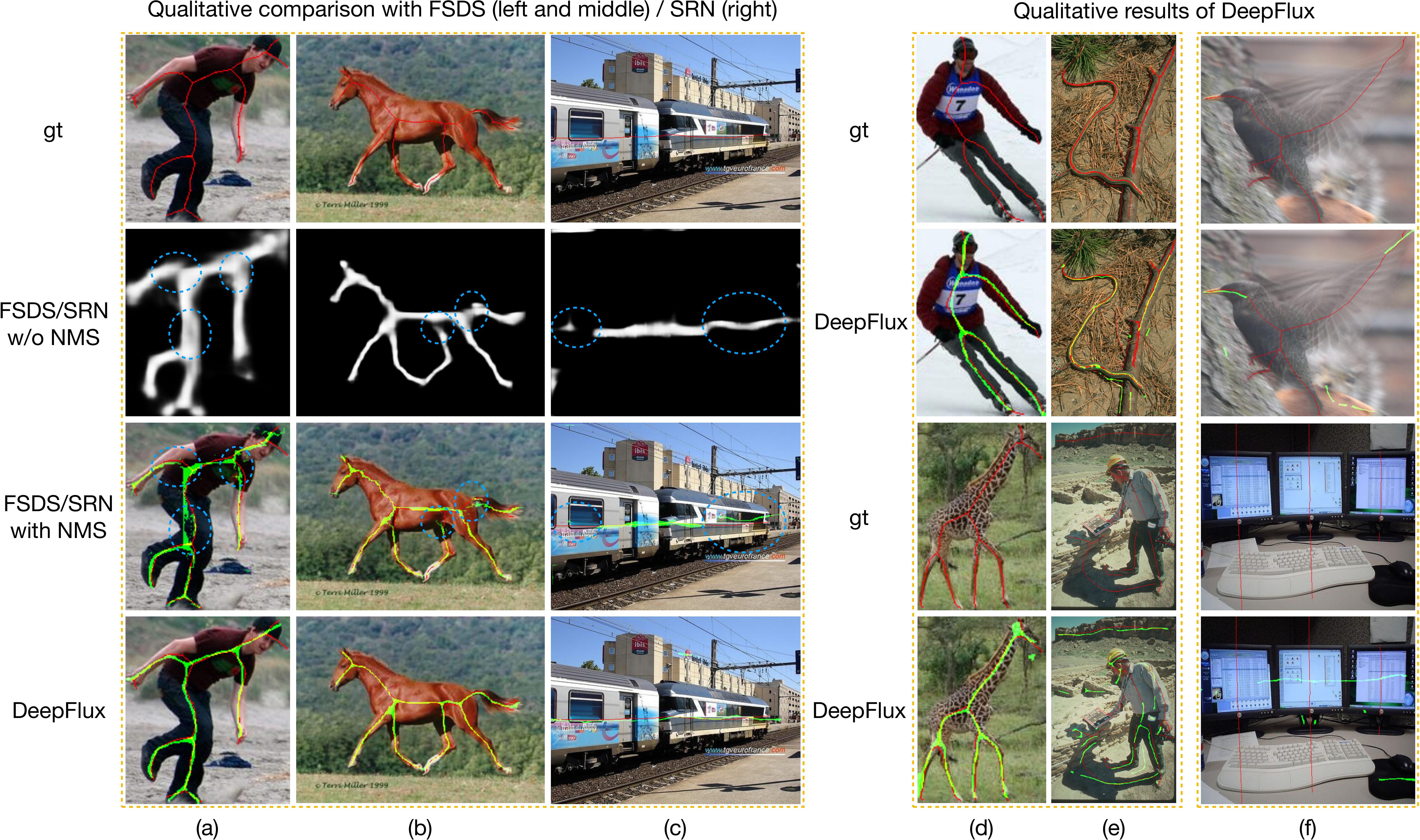}
\vskip 0.2cm
\caption{Qualitative results on SK-LARGE, WH-SYMMAX, and SYM-PASCAL (a-c), SK506 (d), SYMMAX300 (e), and two failure cases (f). Red: GT; Green: detected skeleton; Yellow: detected skeleton and GT overlap.
DeepFlux fails to detect the skeleton on the bird body due the severe blurring. In the second failure example DeepFlux detects a symmetry axis not annotated in the ground truth.
}
\label{fig:qualitativeresults}
\end{figure*}

Similar results are observed for the symmetry detection task.
DeepFlux significantly outperforms \sota\ methods on the SYM-PASCAL dataset, recording an improvement of $4.8\%$ and $7.7\%$ compared to Hi-Fi~\cite{zhao2018hifi} and LSN~\cite{liu2018lsn}, respectively.
On SYMMAX300, DeepFlux also improves over LSN by $1.1\%$.
Some qualitative results are shown in~\reffig{fig:qualitativeresults}, including failure cases.

\subsection{Runtime analysis} \label{sec:runtime}
We decompose runtime analysis into two stages: network inference and post-processing.
Inference on the GPU using VGG16 takes on average \unit[14]{ms} for a $300 \times 200$ image and the post-processing stage requires on average \unit[5]{ms} on the CPU.
As shown in~\reftab{tab:runtime}, DeepFlux is as fast as competing methods while achieving superior performance.

\begin{table}
\begin{center}
\begin{tabular}{|l|c|c|}
\hline
Method & F-measure & Runtime (in \emph{sec}) \\
\hline\hline
HED~\cite{xie2015hed} & 0.497 & 0.014 \\
\hline
FSDS~\cite{shen2016fsds} & 0.633 & 0.017 \\
\hline
LMSDS~\cite{shen2017lmsds} & 0.649 & 0.017 \\
\hline
LSN~\cite{liu2018lsn} & 0.668 & 0.021 \\
\hline
SRN~\cite{ke2017srn} & 0.678 & 0.016 \\
\hline
Hi-Fi~\cite{zhao2018hifi} & 0.724 & 0.030 \\
\hline
DeepFlux (ours) & 0.732 & 0.019 \\
\hline
\end{tabular}
\end{center}
\caption{Runtime and performance on SK-LARGE. For DeeFlux we list the total inference (GPU) + post-processing (CPU) time. }
\label{tab:runtime}
\end{table}

\subsection{Ablation study} \label{sec:ablation}
We study the contribution of the two main modules (ASPP module and flux representation)  to skeleton detection on SK-LARGE and SYM-PASCAL. We first remove the ASPP module and study the effect of the proposed context flux representation compared to a baseline model with the same architecture, but trained for binary classification.
As shown in~\reftab{tab:ablation}, employing a flux representation results in a $2.0\%$  improvement on SK-LARGE and $4.9\%$ on SYM-PASCAL.
We then conduct experiments without using context flux, and study the effect of the increased receptive field offered by the ASPP module.
The ASPP module alone leads to a $1.6\%$ improvement on SK-LARGE and $1.7\%$ on SYM-PASCAL.
This demonstrates that the gains from ASPP and context flux are orthogonal; indeed, combining both improves the baseline model by $\sim4\%$ on SK-LARGE and and $\sim10\%$  on SYM-PASCAL.

We also study the effect of the size of the neighborhood within which context flux is defined.
We conduct experiments with different radii, ranging from $r=3$ to $r=11$, on the SK-LARGE and SYM-PASCAL datasets.
Best results are obtained for $r=7$, and using smaller or larger values seems to slightly decrease performance.
Our understanding is that a narrower context neighborhood provides less contextual information to predict the final skeleton map.
On the other hand, using a wider neighborhood may increase the chance for mistakes in flux prediction around areas of severe discontinuities, such as the areas around boundaries of thin objects that are fully contained in the context neighborhood.
The good news, however, is that DeepFlux is not sensitive to the value of $r$.

Finally, one may argue that simply using a dilated skeleton ground truth is sufficient to make a baseline model more robust in accurately localizing skeleton points.
To examine if this is the case, we retrained our baseline model using binary cross-entropy on the same dilated ground truth we used for DeepFlux.
Without context flux, performance drops to $F=0.673$ (\textcolor{red}{$-6\%$}) on SK-LARGE and to $F=0.425$ (\textcolor{red}{$-8\%$}) on SYM-PASCAL, validating the importance of our proposed representation for accurate localization.

\begin{table}
\begin{center}
\begin{tabular}{|c|c|c|c|}
\hline
Dataset & Context flux & ASPP & F-measure \\
\hline\hline
\multirow{4}{*}{SK-LARGE} & & &0.696 \\
& & \checkmark & 0.712 \\
& \checkmark & & 0.716 \\
& \checkmark & \checkmark & \textbf{0.732} \\
\hline\hline
\multirow{4}{*}{SYM-PASCAL} & & &0.409 \\
& & \checkmark & 0.426\\
& \checkmark & & 0.458\\
& \checkmark & \checkmark & \textbf{0.502} \\
\hline
\end{tabular}
\end{center}
\caption{The effect of the context flux representation and the ASPP module on performance.}
\label{tab:ablation}
\end{table}

\begin{table}
\begin{center}
\begin{tabular}{|c|c|c|c|c|c|}
\hline
Dataset & $r=3$ & $r=5$ & $r=7$ & $r=9$ & $r=11$ \\
\hline \hline
\small{SK-LARGE} & 0.721 & 0.727 & \textbf{0.732} & 0.726 & 0.724\\
\hline
\small{SYM-PASCAL} & 0.481 & 0.498 & \textbf{0.502} & 0.500 & 0.501 \\
\hline
\end{tabular}
\end{center}
\caption{Influence of the context size $r$ on the F-measure.}
\label{tab:diffks}
\end{table}

\section{Conclusion} \label{sec:conclusion}
We have proposed DeepFlux, a novel approach for accurate detection of object skeletons in the wild.
Departing from the usual view of learning-based skeleton detection as a binary classification problem, we have recast it as the regression problem of predicting a 2D vector field of ``context flux''.
We have developed a simple convolutional neural network to compute such a flux, followed by a simple post-processing scheme that can accurately recover object skeletons in $\sim20\emph{ms}.$
Our approach steers clear of many limitations related to poor localization, commonly shared by previous methods, and particularly shines in handling ligature points, and skeletons at large scales.

Experimental results on five popular and challenging benchmarks demonstrate that DeepFlux systematically improves the state-of-the-art, both quantitatively and qualitatively.
Furthermore, DeepFlux goes beyond object skeleton detection, and achieves \sota\ results in detecting generic symmetry in the wild.
In the future, we would like to explore replacing the post-processing step used to recover the skeleton with an appropriate NN module, making the entire pipeline trainable in an end-to-end  fashion.

{\small
\bibliographystyle{ieee}
\bibliography{ref}
}

\end{document}